\documentclass[11pt]{article}
\usepackage[utf8]{inputenc}
\usepackage[T1]{fontenc}
\usepackage[margin=1in]{geometry}
\usepackage{amsmath,amssymb}
\usepackage{booktabs}
\usepackage{array}
\usepackage{tabularx}
\usepackage{graphicx}
\usepackage{placeins}
\usepackage{microtype}
\usepackage[hidelinks]{hyperref}
\usepackage{caption}
\title{Measuring Stability and Failure Behavior in Language Models\\ Under Structured Perturbations}
\author{Samira Golsefid \\ Principal Researcher \\ \texttt{samira@neuroloft.org}}
\date{}

\begin{document}
\maketitle
\begin{abstract}
Language models are usually judged by a single accuracy score, which does not reveal how their performance degrades as inputs are perturbed. We present a graded, multi-family, failure-aware framework for stress-testing reasoning models. It perturbs each problem along a multi-level severity ladder across seven families: six that preserve the answer, paraphrase, input noise, formatting, irrelevant context, context load, and conflicting instructions, and a Knowledge Boundary family that removes answerability so that refusal becomes the correct response. Every test is validity-gated and labeled by its measured severity, and each model is summarized by per-level Accuracy, a magnitude-weighted Stability, and a per-family Collapse Point defined relative to the model's own baseline. Instantiated on the same 100 seed problems used by GSM-Symbolic, expanded into 4,473 gated tests and run on four models spanning capability tiers, the framework exposes structure that an aggregate score hides: the level at which a model fails is family-specific rather than global, and two stressors expose consistent weaknesses across all models: conflicting instructions and questions built on an impossible premise. Recognition of unanswerability is otherwise uneven, reliable on missing information and fabricated evidence but weak on impossible premises. These failure points are invisible to standard accuracy reporting.
\end{abstract}
\section{Introduction}
Language models are overwhelmingly judged by a single number: accuracy on a held-out benchmark. Such a number says little about how a model behaves, whether a trivial rewording flips its answer, how its performance erodes as conditions harden, or how it fails when it fails. A growing body of evidence shows that this scalar is fragile and, on its own, misleading: meaning-preserving changes to a prompt can move accuracy by tens of points [Sclar et al., 2024; Mizrahi et al., 2024], a single irrelevant clause can sharply reduce it [Mirzadeh et al., 2024; Shi et al., 2023], and models can solve familiar problems by pattern-matching in ways that do not survive perturbation [Wu et al., 2024]. Equally important is what a model does when a question cannot be answered at all---whether it abstains or confidently fabricates---a behavior that accuracy on answerable questions cannot capture.

Reliable reasoning requires both: behavioral invariance under meaning-preserving transformations, and appropriate failure---refusing or flagging a question that cannot be answered---when the input leaves the space of valid problems. These concerns are studied across several separate literatures, some that perturb reasoning benchmarks to probe robustness, others that examine failure behavior such as abstention on unanswerable inputs, and they are rarely brought together. As a result, a single question remains open: how does a model behave across the full range and variety of stress it may meet, including problems it should not answer at all? Addressing this calls for a unified, graded, and reusable evaluation.

We close this gap with a framework that stresses a reasoning model along a graded severity ladder across seven perturbation families, paraphrase, input noise, formatting, irrelevant context, context load, conflicting instructions, and Knowledge Boundary testing, and characterizes its failure behavior in the same frame (Section~3.1). Each generated test is validity-gated and labeled by its measured severity rather than the level requested of the generator, separating requested severity from measured severity, and each model is summarized by three quantities: per-level Accuracy, a magnitude-weighted Stability, and a per-family Collapse Point that marks where the model first collapses relative to its own baseline (Section~3.2). We instantiate the framework on the 100 grade-school problems that underlie GSM-Symbolic [Mirzadeh et al., 2024], expanding them into a gated suite of 4,473 admitted tests and evaluating four models that span capability tiers and deployment settings, a reasoning-oriented model (o4-mini), two widely used general-purpose models (gemini-2.5-flash, gpt-4o-mini), and a small open-weight model (phi4-mini).

This finer-grained view exposes structure that an aggregate score hides. Every model degrades under perturbation, but the meaning-preserving surface families are handled robustly by the stronger models, and the level at which a model collapses is family-specific rather than global. Because each model's Collapse Point is measured relative to its own baseline, the comparison holds: gemini-2.5-flash and gpt-4o-mini collapse in very different places, with gpt-4o-mini failing on context-related families that gemini-2.5-flash handles cleanly. The most consistent results are two shared weaknesses: every model reaches its Collapse Point under conflicting instructions, and every model performs poorly on impossible-premise Knowledge Boundary questions, computing an answer instead of flagging the impossibility. The knowledge boundary is otherwise handled unevenly, models reliably refuse questions that omit essential information or cite fabricated evidence but not those resting on an impossible premise. These failures persist even in the strongest model.
In this paper, after reviewing the related work (Section~2), we present:
\begin{itemize}
\item a graded, multi-family, failure-aware framework for stress-testing reasoning models, with three reportable metrics, Accuracy, Stability, and Collapse Point, that turn a single accuracy drop into a degradation curve and a behavioral profile (Section~3);

\item a validity-gated, reusable benchmark built on the GSM-Symbolic seed problems, with a generation-and-gating pipeline that admits only meaning-preserving, or for Knowledge Boundary, genuinely unanswerable, tests and labels each by its measured severity (Section~4);

\item an empirical study of four models across capability tiers, including an open-weight model, that reveals a consistent failure pattern, surface-robust yet uniformly fragile under conflicting instructions and at the impossible-premise knowledge boundary, and relates it to prior single-axis results on the same problems (Section~5).
\end{itemize}

\section{Related Work}
Prior work spans three threads: evaluation beyond accuracy, reasoning robustness under input perturbation, and model behavior at the knowledge boundary.

CheckList [Ribeiro et al., 2020], titled "Beyond Accuracy," proposes behavioral testing, including invariance tests that apply label-preserving perturbations and expect unchanged predictions. HELM [Liang et al., 2022] reports seven metrics, accuracy, calibration, robustness, fairness, bias, toxicity, and efficiency, across sixteen scenarios. Both, however, target classification or aggregate model comparison, not how a reasoning model degrades under intensifying stress. Other studies show that the scalar is not even reproducible: few-shot example order [Lu et al., 2022] and example selection [Zhao et al., 2021] shift it, and meaning-preserving format changes alone can move accuracy by up to 76 points on a single task [Sclar et al., 2024], motivating calls to report a range rather than a point [Mizrahi et al., 2024; Alzahrani et al., 2024; Zhu et al., 2023]. The closest prior work, SCORE [Nalbandyan et al., 2025], reports accuracy ranges and a consistency rate, but quantifies how much the number moves rather than how the model fails, perturbing the prompt wrapper rather than the reasoning problem itself.

A second body of research perturbs benchmark inputs and measures the effect on accuracy. Most of this work centers on grade-school math (GSM8K) [Cobbe et al., 2021]. GSM-Symbolic [Mirzadeh et al., 2024] regenerates problems from templates and shows that a single seemingly-relevant clause can cut accuracy by up to 65\%; GSM-Plus [Li et al., 2024] applies eight perturbation types across five perspectives; GSM-IC [Shi et al., 2023] inserts irrelevant sentences; GSM-Hard [Gao et al., 2023] enlarges numbers; and 
R-GSM [Chen et al., 2024] reorders logically-independent premises, with permutation dropping accuracy by over 30\% on deductive reasoning. Beyond GSM8K, MATH-Perturb [Huang et al., 2025] contrasts simple and hard perturbations of competition problems, and contamination-controlled re-tests [Zhang et al., 2024] attribute part of the drop to memorization. Together, these results suggest models often recite procedures rather than reason robustly [Wu et al., 2024].

Most of this work inherits the limitation it set out to expose. Each applies one or a few perturbation types in isolation; each applies them at a single intensity, with the partial exception of GSM-Symbolic's clause count, yielding a point rather than a degradation curve (CheckList's invariance tests are likewise binary); and each collapses its findings into another scalar, an accuracy drop, that reports how much performance fell but not how the model failed. The closest graded designs remain single-axis: MATH-Perturb uses two intensities [Huang et al., 2025], and a numeric-sensitivity study ramps a single perturbation type, reporting up to a 51.55\% drop [Sun et al., 2025]. The nearest concurrent work is MedDialBench [Luo et al., 2026], which adds graded severity and dose-response profiling but for multi-turn medical dialogue rather than single-turn reasoning, and the Robust Reasoning Benchmark [Golikov et al., 2026], which organizes perturbations into families with failure-mode analysis but at a single intensity and with adversarial-encoding transforms (ciphers, reversals) rather than naturalistic variation. To the best of our knowledge, no prior evaluation combines a graded severity ladder, multiple naturalistic perturbation families, and failure-behavior characterization for single-turn reasoning on a validity-gated, reusable benchmark.

A third line of research studies how a model should fail, a question studied apart from robustness. Abstention on unanswerable inputs dates to SQuAD 2.0 [Rajpurkar et al., 2018]. False-premise questions are isolated by FalseQA [Hu et al., 2023] and catalogued by KUQ [Amayuelas et al., 2023], whose taxonomy of unanswerability, future events, false assumptions, and open problems, mirrors our Knowledge Boundary family. Whether models know what they know is a calibration question [Kadavath et al., 2022], and confident falsehoods are measured by TruthfulQA [Lin et al., 2022]. Under pressure, models exhibit sycophancy [Sharma et al., 2023] and inconsistent handling of conflicting instructions [Wallace et al., 2024; Geng et al., 2025]. We connect these literatures by measuring refusal on the same severity ladder as every other stressor, so that robustness and failure behavior are reported in one frame rather than in two separate evaluations.

\section{Framework}
In this section, we present our framework for evaluating reasoning robustness: the perturbation families and the severity ladder along which each is graded (Section 3.1), and the per-test and aggregate metrics used to score behavior under them (Section 3.2). A family defines a type of stress, while the severity level defines its intensity. Every generated test is admitted only after passing the family's acceptance gate.

\subsection{Perturbation families}
Our framework organizes input perturbations into seven families, each a distinct type of stress a competent reasoner should withstand, graded along a severity ladder of increasing dose (five levels throughout; see Section~4.2). The six meaning-preserving families fall into three groups by what they perturb: surface form (Semantic Variation, Input Quality, Structural/Format), surrounding context (Context Interference, Context Load), and task instructions (Conflict Instruction Stress). The seventh, Knowledge Boundary, is the deliberate exception: it perturbs answerability itself. Every meaning-preserving test must preserve the original numbers and gold answer and pass its family's validity gate before admission; Knowledge Boundary instead must pass an epistemic-validity gate confirming the question is genuinely unanswerable. Each admitted test is then classified by its measured severity (Section 3.2) rather than the requested level. This separates the severity we intended to apply from the severity actually achieved: a candidate generated at L3 but measured as L2 is recorded as L2, so every degradation curve is plotted against achieved rather than requested severity.

Meaning preservation is enforced by two complementary checks. An LLM entailment (NLI) gate---the generated text must entail the original---is applied to Semantic Variation, Input Quality, Context Interference, and Context Load. Structural/Format needs no NLI gate: its format-only edits leave every word and number in place, so the structural gate alone suffices. An LLM judge additionally verifies meaning preservation, confirming that Semantic Variation and Input Quality leave the problem's content and answer intact, and that the inserted text is irrelevant for Context Interference, non-essential for Context Load, 
or that the generated conflict preserves the original meaning for Conflict Instruction Stress, where this judge is the sole meaning check. Knowledge Boundary instead uses an epistemic-validity judge that confirms the question is genuinely unanswerable. Table 1 summarizes the families; the gates are detailed in Section~4.2.

\textbf{Semantic Variation (SV).} SV paraphrases the problem, preserving meaning and every quantity; a model that reasons over content should be invariant to rewording, and sensitivity signals reliance on lexical surface form (SEARs [Ribeiro et al., 2018]; see also [Wang \& Zhao, 2024]). Higher levels paraphrase more aggressively, lowering lexical similarity to the original while holding the numbers and answer fixed.

\textbf{Input Quality (IQ).} IQ injects noise, misspellings and character- or word-level corruption, without altering content, mimicking the degraded input models meet in practice [Wang et al., 2023; Dong et al., 2023]. Higher levels raise the corruption density.

\textbf{Structural/Format (F).} F changes how a problem is presented, layout, delimiters, and formatting, while preserving the exact words, isolating whether the answer depends on incidental form; trivial formatting changes are a documented source of large accuracy swings [Sclar et al., 2024]. Higher levels apply more formatting disruption, with the exact words preserved.

\textbf{Context Interference (CI).} CI inserts irrelevant but topically plausible sentences that leave the question unchanged, probing whether a model can ignore information it does not need [Shi et al., 2023]. Higher levels add more, and less distinguishable, distracting content.

\textbf{Context Load (CL).} CL expands the context with on-topic but non-essential material, increasing the volume a model must process; reasoning degrades as input grows, even below the context limit, and mid-context information is often lost [Liu et al., 2024; Levy et al., 2024]. Higher levels increase the expansion (admission requires the perturbed-to-original length ratio to reach the level's band).

\textbf{Conflict Instruction Stress (CIS).} CIS leaves the problem and its gold answer untouched but appends contradictory or adversarial directives, testing whether the model holds the correct task under pressure [Wallace et al., 2024; Geng et al., 2025]. Higher levels escalate the conflict's explicitness along a five-tier conflict-explicitness score (CES), set by the strongest directive detected in the added sentences: a mild suggestion (tier 1; e.g. "perhaps", "consider"), a wrong-operation hint (2), an explicit override (3; e.g. "ignore", "disregard"), an authority command (4; e.g. "System:", "you must"), and a deep-hierarchy attack (5; e.g. "Administrator:", "override").

\textbf{Knowledge Boundary (KB).} KB is the one family that deliberately changes what a problem asks: it renders the question unanswerable, so the correct behavior is to refuse or flag it rather than produce a number. A trustworthy model should recognize the limits of what it can answer [Rajpurkar et al., 2018; Hu et al., 2023; Amayuelas et al., 2023]; the failure this family is designed to surface is producing a confident answer to a question that has none. Unlike the other families, KB's five levels are not a perturbation dose but five qualitatively different boundary types: missing essential information (L1), fabricated evidence (L2), a not-yet-knowable value (L3), a fabricated entity (L4), and an impossible premise (L5). We index them L1--L5 for consistency with the framework; the ordering reflects the boundary taxonomy used in this study rather than a measured surface-severity scale.
Five families place a test on the ladder using a deterministic surface metric (Table 2); the other two are not surface-based: Conflict Instruction Stress uses a five-tier conflict-explicitness rubric, and Knowledge Boundary's levels are categorical, as described above.

\begin{table}[t]\centering
\caption{Perturbation families and the robustness property each one stresses.}
\footnotesize
\begin{tabularx}{\textwidth}{*{4}{>{\raggedright\arraybackslash}X}}\toprule
Family & Robustness property stressed & Levels escalate by & Expected stable behavior \\ \midrule
Semantic Variation & invariance to rewording & more aggressive paraphrase & same answer \\
Input Quality & tolerance to noisy input & denser corruption & same answer \\
Structural/Format & independence from formatting & more format disruption & same answer \\
Context Interference & ignoring irrelevant context & more / less-distinguishable distractors & same answer \\
Context Load & robustness to context load & more non-essential padding & same answer \\
Conflict Instruction Stress & holding the task under conflict & stronger, more explicit conflict & same answer \\
Knowledge Boundary & recognizing unanswerability & qualitatively different boundary, not dose & refuse / flag \\
\bottomrule\end{tabularx}
\end{table}

\subsection{Metrics}
We score each test on two per-test axes, its output correctness and the magnitude of its perturbation, and aggregate either over any set of tests S (a family, a severity level, a family $\times$ level cell, the unperturbed baseline, or all data) into three quantities: Accuracy, Magnitude, and Stability. Below, i indexes a single test, |S| is the number of tests in S, $\Sigma$$_i$ sums over i $\in$ S, and O and G denote the original and perturbed problem text. Per-test quantities. For each test i we compute two scores in [0, 1].

Output correctness (sim$_i$). Agreement between the model output and the gold answer. For the six answer-preserving (numeric) families,

\[ \mathrm{sim}_i = 1 \ \text{if}\ |p_i - g_i|/|g_i| < 10^{-3},\ \text{and } 0 \text{ otherwise}\quad (\mathrm{sim}_i\in\{0,1\}), \]
where p$_i$ is the model's predicted answer and g$_i$ = y*$_i$ is the gold answer (exact match within a 0.1\% relative tolerance). For Knowledge Boundary, correctness is scored behaviorally by an LLM judge rather than by answer matching: sim$_i$ rates how well the response recognizes that the question cannot be answered, on a four-point rubric sim$_i$ $\in$ \{0, 0.25, 0.75, 1.0\}. A score of 1.0 explicitly rejects or flags the unanswerable question (e.g.\ names the impossible premise or asks for the missing quantity); 0.75 signals uncertainty but stops short of rejection; 0.25 stays on topic yet never addresses the boundary, neither flagging the problem nor committing to an answer; and 0 marks the core failure this family targets---a confident answer that treats the impossible premise as legitimate.

Perturbation magnitude (score$_i$). How much the perturbation changed the input text, as a single unified surface-change measure that is identical for every family:

score$_i$ = 1 - SIM$_i$,

where SIM$_i$ = 2M / T is the character-sequence similarity between the original O and the generated G, with M the number of matching characters and T = len(O) + len(G) their combined length (computed as SequenceMatcher(O, G).ratio()); so score$_i$ = 1 - SIM$_i$ is the fraction of the character sequence that changed.

Aggregate metrics. Each axis is averaged over a set S of tests; we report three quantities.

Accuracy. Mean output correctness over S:

\[ \mathrm{Accuracy}(S) = \frac{1}{|S|}\sum_{i\in S} \mathrm{sim}_i. \]
On the unperturbed baseline this is the baseline accuracy acc(L$_0$); on the tests at level k of a family it is acc(L$_k$), the per-level point whose plot against k is that family's degradation curve.

Magnitude. Mean perturbation magnitude over S:

\[ \mathrm{Magnitude}(S) = \frac{1}{|S|}\sum_{i\in S} \mathrm{score}_i. \]

Stability. Output correctness weighted by perturbation magnitude, so harder (more perturbed) tests count more. Stability gives more weight to tests where the perturbation was harder, while Accuracy gives equal weight to every test:

\[ \mathrm{Stability}(S) = \frac{\sum_{i\in S} \mathrm{sim}_i\cdot \mathrm{score}_i}{\sum_{i\in S}\mathrm{score}_i}. \]
For example, a test with $\mathrm{SIM}=0.6$ has $\mathrm{score}=0.4$, so answering it correctly adds $0.4$ to both the numerator and the denominator, whereas a near-unperturbed test ($\mathrm{score}\approx 0$) barely counts; Stability is thus weighted toward the largest-magnitude perturbations. A model that succeeds on low-magnitude perturbations but fails on high-magnitude ones will have Accuracy \textgreater{} Stability; a model whose failures are spread evenly across magnitudes will have Accuracy $\approx$ Stability.

Collapse Point (CP). The lowest severity level at which per-level accuracy drops below a fixed fraction of the model's own baseline:

\[ \mathrm{CP} = \min\{\,k : \mathrm{acc}(L_k) < \tau\,\},\qquad \tau = 0.80\times \mathrm{acc}(L_0). \]
Defining $\tau$ relative to each model's baseline, rather than as an absolute cutoff, makes the Collapse Point comparable across models of differing overall ability. In this study, we use 80\% as a moderate-failure threshold. CP is undefined when accuracy never falls below $\tau$.

Severity metrics (level assignment). The same score$_i$ that gives the magnitude also places most tests on the ladder: Semantic Variation, Input Quality, Structural/Format, Context Interference, and Context Load are assigned the level whose family-specific band contains score$_i$ (Table 2). Conflict Instruction Stress is placed by a conflict-explicitness score (CES) and Knowledge Boundary is categorical. We use deterministic surface metrics rather than embeddings so level assignment is reproducible, and enforce meaning-preservation separately at the generation gate (Section~4.2).

\begin{table}[t]\centering
\caption{Metric used to place each family on the L1--L5 severity ladder. Five families are placed by binning the perturbation magnitude (Section~3.2) against family-specific per-level bands; Conflict Instruction Stress uses a conflict-explicitness score (CES) and Knowledge Boundary is categorical.}
\footnotesize
\begin{tabularx}{\textwidth}{*{4}{>{\raggedright\arraybackslash}X}}\toprule
Family & Level metric & Range & What rises with level \\ \midrule
Semantic Variation & score & [0, 1] & more aggressive paraphrase \\
Input Quality & score & [0, 1] & denser noise / corruption \\
Structural/Format & score & [0, 1] & more format disruption \\
Context Interference & score & [0, 1] & more added-content load \\
Context Load & score; the gate additionally requires a word-expansion ratio words(G) / words(O) in band & [0, 1] & more context expansion \\
Conflict Instruction Stress & CES $\in$ \{1, \dots{}, 5\}, normalized as (CES - 1) / 4 & [0, 1] & stronger, more explicit conflict \\
Knowledge Boundary & none (categorical) & - & qualitatively different boundary type \\
\bottomrule\end{tabularx}
\end{table}

\section{Experiment}
This section puts the framework of Section~3 into practice. We evaluate language models on a single-turn grade-school math task, holding the underlying problems fixed while subjecting them to graded, multi-family perturbations, and measuring not only how far accuracy falls but how each model fails. We describe the seed dataset (Section~4.1), the pipeline that generates and validity-gates the test suite (Section~4.2), and the systems under test and the protocol for running and scoring them (Section~4.3); all quantities follow Section~3.2.

\subsection{Dataset}
We build on grade-school math word problems, a domain where each problem has a single, unambiguous numerical answer. Correctness can therefore be scored exactly, and any change in accuracy under a meaning-preserving perturbation reflects fragility rather than disagreement about the gold answer.

Our seeds are the same 100 grade-school problems used by GSM-Symbolic [Mirzadeh et al., 2024], a subset of the GSM8K test split [Cobbe et al., 2021], publicly available via apple/GSM-Symbolic on HuggingFace. We retain each problem's original GSM8K index and apply our perturbation families (Section 3) to the original problems rather than to symbolic templates, using the GSM-Symbolic seed set with our own stress-testing methodology.

Each seed carries a verified numerical ground-truth answer and plays two roles: it is the unperturbed item against which baseline accuracy acc(L0) is measured, and the source from which all of its perturbed variants are generated. The choice of 100 seeds is a deliberate trade-off: expanded across seven families and five levels and then validity-gated, the seeds yield thousands of admitted tests, keeping per-family sample sizes adequate while bounding compute cost; fixing the set at exactly the 100 GSM-Symbolic seeds also lets us compare directly against prior work on the same problems (Section~5.3).

\subsection{Generating the test suite}
From each seed problem we generate candidate tests across the seven perturbation families and five severity levels defined in Section~3.1. Candidates are produced by an LLM-based generator under family-specific instructions, with multiple attempts and a range of sampling temperatures so that enough candidates clear each level's severity band.

Every candidate passes through a sequence of automatic validity gates. A structural gate checks that the answer-bearing quantities are preserved; a meaning-preservation gate---an LLM-based textual-entailment (NLI) check that labels the candidate entailment, neutral, or contradiction against the original---is applied to Semantic Variation, Input Quality, Context Interference, and Context Load; Structural/Format is exempt, since its format-only edits leave every word and number intact. An LLM judge further checks that Semantic Variation and Input Quality preserve meaning, and that the material added in Context Interference, Context Load, and Conflict Instruction Stress is respectively irrelevant, non-essential, or a genuine conflict. Knowledge Boundary is gated differently: an epistemic-validity judge confirms the question has become genuinely unanswerable; its five categorical boundary types are defined in Section~3.1.

Crucially, each admitted test is labelled by its measured severity (Section~3.2) rather than the level we requested, separating the intensity we asked for from the intensity we obtained.

Across the 100 seeds the generator produced 5,673 candidates, of which 4,473 (78.9\%) passed all gates and were admitted; the remaining 1,200 were rejected. Table 3 reports admitted counts per family. Per-level bands are given in Table 2; the exact gate thresholds and judge prompts are in the code repository.

\subsection{Running the LLMs}
We evaluate four systems under test (SUTs): o4-mini (OpenAI), gemini-2.5-flash (Google), gpt-4o-mini (OpenAI), and phi4-mini (Microsoft, run locally via Ollama). The set spans a reasoning-oriented model, two widely used general-purpose models, and a small open-weight model, letting us compare degradation behavior across capability tiers and deployment settings.

Each SUT is given the identical test content and answers every admitted test plus the unperturbed baseline. We query zero-shot instructing each model to return its final numerical answer in a machine-readable format; a single tolerant parser recovers the answer, accepting either a JSON object or a final ``\#\#\#\# \textless{}number\textgreater{}'' line and falling back to the last number. Decoding is deterministic (temperature 0) wherever the API exposes it; the reasoning model o4-mini does not permit decoding controls and uses its defaults. Runs are resumable per seed. For the baseline we query each seed three times unperturbed (300 evaluations) and take the mean as acc(L0), anchoring each model's Collapse Point to its own baseline (Section~3.2).

o4-mini and gemini-2.5-flash returned the final answer directly, whereas gpt-4o-mini and phi4-mini produced an explicit step-by-step solution; in each case we take the elicited response as the output. The full prompt given to each model is in Appendix~A.

Responses are scored as in Section~3.2: for the six answer-preserving families a response is correct when its parsed numeric answer matches the gold answer within a 0.1\% relative tolerance, and for Knowledge Boundary a gpt-4o-mini judge scores each response on a four-point epistemic rubric (1.0/0.75/0.25/0). Each model thus produces 4,773 scored responses, 4,473 perturbed tests and 300 baseline evaluations, from which we compute the per-family Accuracy, Magnitude, Stability, and Collapse Point, compared across models in Section~5.

\begin{table}[t]\centering
\caption{Admitted (validity-gated) tests per perturbation family across the 100 seeds.}
\footnotesize
\begin{tabular}{ll}\toprule
Family & Admitted tests \\ \midrule
Structural/Format (F) & 818 \\
Input Quality (IQ) & 751 \\
Context Load (CL) & 737 \\
Context Interference (CI) & 605 \\
Conflict Instruction Stress (CIS) & 600 \\
Semantic Variation (SV) & 573 \\
Knowledge Boundary (KB) & 389 \\
Total & 4,473 \\
\bottomrule\end{tabular}
\end{table}

\section{Results}
We evaluate the four models of Section~4 on the 4,473-test suite plus the unperturbed baseline, reporting the metrics of Section~3.2. We compare the models across the severity ladder (Section~5.1), then turn to per-family stability, collapse points, and behavioral patterns, including the conflict-instruction and Knowledge Boundary failures common to all models (Section~5.2), and finally relate the findings to prior work on the same problems (Section~5.3).

\subsection{Accuracy across models and severity levels}
Figure 1 plots aggregate accuracy against measured severity. Every model degrades: from baseline (L0) to L5, accuracy falls from 0.98 to 0.65 (o4-mini), 0.94 to 0.46 (gemini-2.5-flash), 0.89 to 0.41 (gpt-4o-mini), and 0.72 to 0.41 (phi4-mini). The decline is uneven: accuracy holds up comparatively well through L4 and then collapses at L5, the level that carries the impossible-premise Knowledge Boundary tests, on which every model fails almost completely (Section~5.3). Through L1--L4 the capability ordering is preserved and degradation is gentle, so models tolerate mild and moderate perturbation; the heavy loss is concentrated at the impossible-premise boundary, where even the strongest model drops to 0.65 and the others fall to 0.41--0.46.

\begin{figure}[t]\centering
\includegraphics[width=0.7\linewidth]{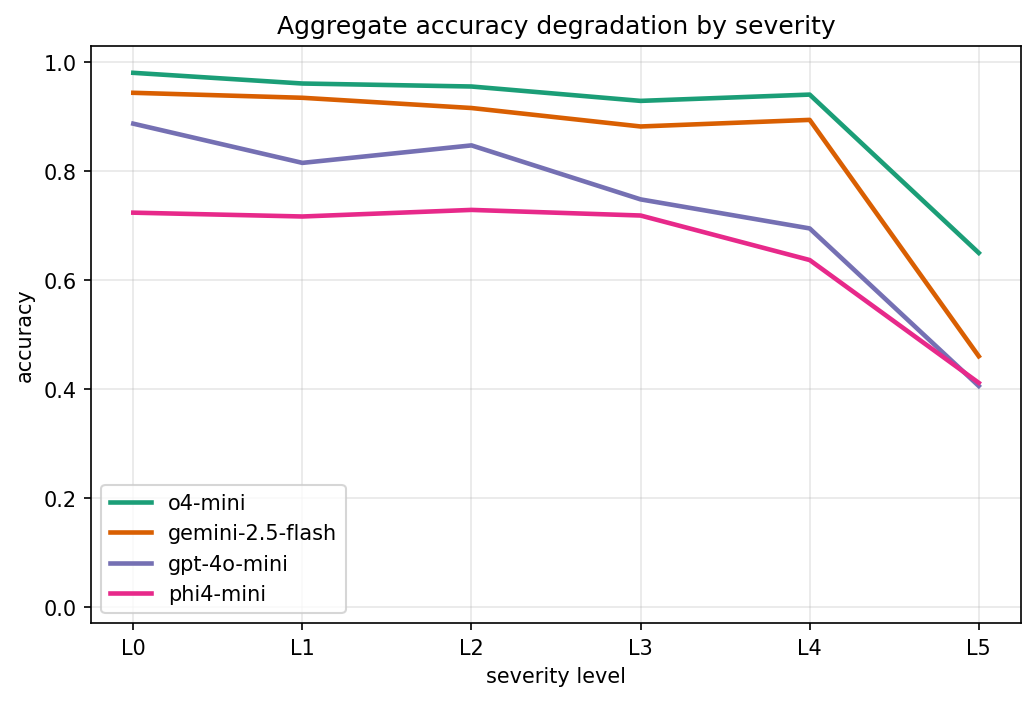}
\caption{Aggregate accuracy by measured severity level (L0 = unperturbed baseline), per model.}
\end{figure}

\subsection{Stability, collapse points, and behavioral patterns}
Stability (Table 4) weights each response by its perturbation magnitude (Section~3.2), rewarding models that stay correct where the perturbation is strongest. Magnitude-weighting preserves the baseline model ordering (o4-mini $>$ gemini-2.5-flash $>$ gpt-4o-mini $>$ phi4-mini); its value lies in the per-family structure it exposes, which a single aggregate score conceals. That structure is sharp: the meaning-preserving surface families (Semantic Variation, Structural/Format, Input Quality) are handled robustly by the stronger models (stability $\ge$ 0.96 for o4-mini and $\ge$ 0.91 for gemini-2.5-flash), whereas gpt-4o-mini falls off on the context families (Context Interference 0.68, Context Load 0.65) and phi4-mini on Input Quality (0.47). Knowledge Boundary and Conflict Instruction Stress are the two lowest-stability families: Knowledge Boundary ranges 0.66--0.77 (the single lowest family for o4-mini and gemini-2.5-flash) and Conflict Instruction Stress 0.56--0.87.

\begin{table}[t]\centering
\caption{Per-family Stability (magnitude-weighted output similarity, Section~3.2) by model. The Baseline row reports the mean output similarity on unperturbed problems; all other rows are magnitude-weighted Stability.}
\footnotesize
\begin{tabular}{lllll}\toprule
Category & o4-mini & gemini-2.5-flash & gpt-4o-mini & phi4-mini \\ \midrule
Baseline & 0.980 & 0.943 & 0.893 & 0.723 \\
Semantic Variation & 0.965 & 0.917 & 0.867 & 0.685 \\
Input Quality & 0.989 & 0.935 & 0.853 & 0.472 \\
Structural/Format & 0.985 & 0.946 & 0.812 & 0.732 \\
Context Interference & 0.966 & 0.947 & 0.683 & 0.729 \\
Context Load & 0.990 & 0.934 & 0.648 & 0.776 \\
Conflict Instruction Stress & 0.871 & 0.740 & 0.692 & 0.561 \\
Knowledge Boundary & 0.695 & 0.662 & 0.709 & 0.768 \\
\bottomrule\end{tabular}
\end{table}

Table 5 locates each model's Collapse Point, the first severity level at which a family's accuracy drops below 80\% of that model's baseline (``---'' = no collapse through L5). Every model collapses under Conflict Instruction Stress (L5 for o4-mini and gemini-2.5-flash, L2 for gpt-4o-mini, L4 for phi4-mini), and Knowledge Boundary collapses for all four as well, though at different levels---L3 (the not-yet-knowable boundary) for the three API models and L5 (impossible premise) for phi4-mini---reflecting that recognition of unanswerability degrades across several boundary types rather than failing at a single one (Section~5.3). o4-mini and gemini-2.5-flash hold every meaning-preserving family through L5, breaking only on these two. gpt-4o-mini is markedly more fragile, additionally collapsing on Context Interference (L1) and Context Load (L3); phi4-mini additionally collapses on Input Quality (L2). Conflict Instruction Stress and the knowledge boundary---most acutely the impossible-premise level (L5), where every model fails (Section~5.3)---are thus the two systemic weak points shared across models.

\begin{table}[t]\centering
\caption{Collapse Point per family and model: the lowest severity level $k$ with acc(L$_k$) $<$ 0.80$\cdot$acc(L$_0$). ``---'' means the family never falls below the threshold through L5.}
\footnotesize
\begin{tabular}{lllll}\toprule
Family & o4-mini & gemini-2.5-flash & gpt-4o-mini & phi4-mini \\ \midrule
Semantic Variation & --- & --- & --- & --- \\
Input Quality & --- & --- & --- & L2 \\
Structural/Format & --- & --- & --- & --- \\
Context Interference & --- & --- & L1 & --- \\
Context Load & --- & --- & L3 & --- \\
Conflict Instruction Stress & L5 & L5 & L2 & L4 \\
Knowledge Boundary & L3 & L3 & L3 & L5 \\
\bottomrule\end{tabular}
\end{table}

Recognition of unanswerability is uneven across the five boundary types---here accuracy is the fraction of a type's unanswerable questions a model correctly refuses or flags, as a range across the four models. Models are most reliable on missing information (L1, 0.92--0.95) and fabricated evidence (L2, 0.78--0.90), correctly declining to answer. Recognition weakens on not-yet-knowable values (L3, 0.36--0.74) and fabricated entities (L4, 0.33--0.67), and collapses on impossible premises (L5, 0.07--0.08 for the three API models, 0.47 for phi4-mini): instead of flagging the impossibility, the model carries the arithmetic through and returns a number. Because L5 carries these impossible-premise tests, it is where aggregate accuracy falls off the cliff in Figure 1. Aggregated over all five types, Knowledge Boundary accuracy is 0.55--0.73, far from the total failure the L5 drop alone would suggest; the impossible-premise gap is shared by every model and unclosed by capability---the strongest reasoning model (o4-mini) does no better than gpt-4o-mini.

Figure 2 shows the same story visually, placing each family in the magnitude--stability plane partitioned into four quadrants: stable (low magnitude, high stability), robust (high magnitude, high stability), unstable (low magnitude, low stability), and failure (high magnitude, low stability). For all four models the meaning-preserving surface families cluster in the stable/robust band; Conflict Instruction Stress drifts toward unstable, most for gpt-4o-mini and phi4-mini; and Knowledge Boundary sits at high magnitude with output stability below the threshold, in the failure region, pulled down by the impossible-premise boundary even though the other boundary types are handled well.

\subsection{Relation to prior results on GSM8K / GSM-Symbolic}
Our seeds are exactly the 100 problems GSM-Symbolic is built from (Section~4.1), letting us position these results against prior work on the same data. Methodologically, earlier studies on this set apply a single perturbation type at one intensity and report one accuracy-drop, often a large one on the models they evaluated: GSM-IC inserts irrelevant sentences [Shi et al., 2023], GSM-Plus applies eight perturbation types [Li et al., 2024], R-GSM reorders premises, which drops accuracy by over 30\% [Chen et al., 2024], and GSM-Symbolic varies a clause count, where a single added clause cuts accuracy by up to 65\% [Mirzadeh et al., 2024]. We instead run seven families on a graded ladder over the same problems and add a failure-behavior family, turning a point estimate into a degradation curve and a behavioral profile. Qualitatively our families reproduce these findings while refining them: the distraction effect of GSM-IC [Shi et al., 2023] appears in Context Interference, but only gpt-4o-mini collapses under it; the long-context degradation of Liu et al. [2024] and Levy et al. [2024] appears as gpt-4o-mini's Context Load collapse; and the format and paraphrase fragility of Sclar et al. [2024], Ribeiro et al. [2018] and Wang \& Zhao [2024] is largely absorbed by the stronger models in Structural/Format and Semantic Variation. Most importantly, the abstention failures isolated by FalseQA [Hu et al., 2023], KUQ [Amayuelas et al., 2023] and SQuAD 2.0 [Rajpurkar et al., 2018] reappear here, on the same problems and under the same ladder, in the Knowledge Boundary collapse reported above, sharpest on impossible-premise questions. Because o4-mini and gemini-2.5-flash postdate these benchmarks and our prompting differs, we read these correspondences qualitatively, not numerically.

\begin{figure}[htbp]\centering
\includegraphics[width=1\linewidth]{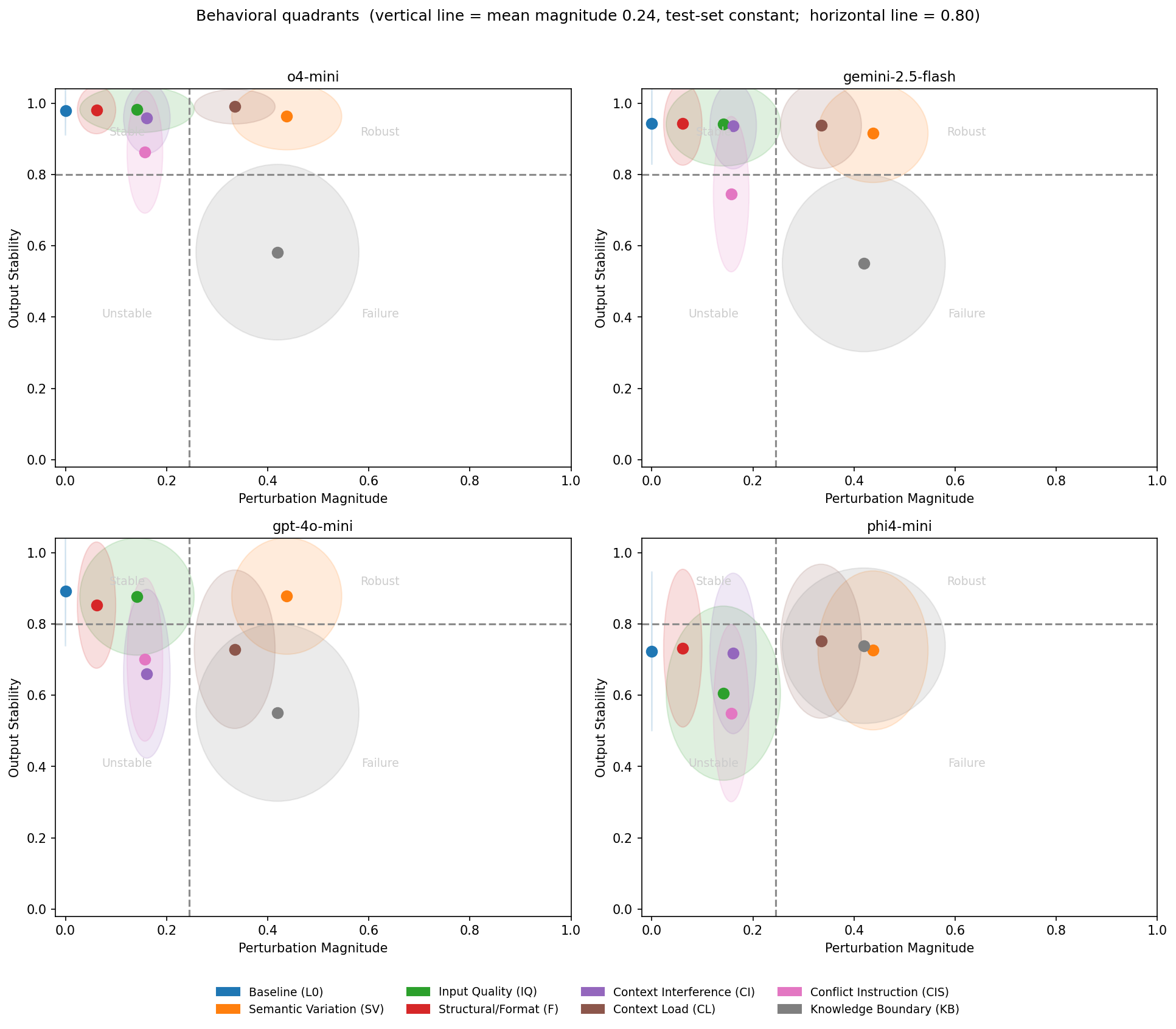}
\caption{Behavioral quadrants per model: each family at its mean perturbation magnitude (x) and output stability (y). The dashed lines mark the stability threshold (0.80) and the mean perturbation magnitude---a property of the test set, identical across models.}
\end{figure}
\FloatBarrier

\section{Discussion and Conclusion}
Taken together, the results support the framework's central claim: a graded, multi-family, failure-aware evaluation reveals structure an aggregate accuracy number cannot. The same model can be robust to heavy surface perturbation yet break under conflicting instructions and on impossible-premise questions; two models can have very different collapse profiles (gemini-2.5-flash and gpt-4o-mini), a contrast that holds because the Collapse Point is measured relative to each model's own baseline; and the level at which a model fails is family-specific rather than global. The recurring weak points, holding the task under conflicting instructions and recognizing impossible premises, are precisely the behaviors that matter for deployment yet are invisible to standard accuracy reporting.

Several limitations qualify these findings. The study uses one task family (grade-school math) on 100 seed problems; broader domains and larger seed sets are needed to test generality. The generation and validity-gating pipeline relies in part on LLM-based entailment and judging, and Knowledge Boundary response scoring uses an LLM judge with a fixed four-point rubric (one level of which was never assigned in practice). These components may introduce evaluator error or systematic judge bias; future work should validate a sample against human annotations and assess inter-judge agreement.

We introduced a graded, multi-family, failure-aware framework for measuring how language models behave under structured perturbation, instantiated on the 100 GSM-Symbolic seed problems as a validity-gated, reusable suite. Its degradation curves, magnitude-weighted stability, per-family Collapse Points, and behavioral quadrants are designed to outlast any single model: the absolute numbers will shift as models improve, but the diagnostic structure remains informative. Natural extensions include further task domains, larger seed sets, uniform multi-model evaluation protocols, and more granular severity ladders---finer bands for the surface families and additional boundary types for Knowledge Boundary---to resolve where each model fails more precisely.

\textbf{Code and data availability.} All code, prompts, per-level bands, gate thresholds, and the generated test suite are available at https://github.com/neuroloft-ai/stability.

\section*{References}
\begingroup\setlength{\parindent}{0pt}\setlength{\hangindent}{1.2em}\small
Alzahrani, N., Alyahya, H. A., Alnumay, Y., AlRashed, S., Alsubaie, S., Almushayqih, Y., Mirza, F., Alotaibi, N., Al-Twairesh, N., Alowisheq, A., Bari, M. S., \& Khan, H. (2024). When Benchmarks are Targets: Revealing the Sensitivity of Large Language Model Leaderboards. ACL 2024, 13787--13805. arXiv:2402.01781.\par\smallskip
Amayuelas, A., Wong, K., Pan, L., Chen, W., \& Wang, W. Y. (2023). Knowledge of Knowledge: Exploring Known-Unknowns Uncertainty with Large Language Models. Findings of ACL 2024. arXiv:2305.13712.\par\smallskip
Chen, X., Chi, R. A., Wang, X., \& Zhou, D. (2024). Premise Order Matters in Reasoning with Large Language Models. ICML 2024. arXiv:2402.08939.\par\smallskip
Cobbe, K., Kosaraju, V., Bavarian, M., Chen, M., Jun, H., Kaiser, L., Plappert, M., Tworek, J., Hilton, J., Nakano, R., Hesse, C., \& Schulman, J. (2021). Training Verifiers to Solve Math Word Problems. arXiv:2110.14168.\par\smallskip
Dong, G., Zhao, J., Hui, T., Guo, D., Wan, W., Feng, B., Qiu, Y., Gongque, Z., He, K., Wang, Z., \& Xu, W. (2023). Revisit Input Perturbation Problems for LLMs: A Unified Robustness Evaluation Framework for Noisy Slot Filling Task. NLPCC 2023. arXiv:2310.06504.\par\smallskip
Gao, L., Madaan, A., Zhou, S., Alon, U., Liu, P., Yang, Y., Callan, J., \& Neubig, G. (2023). PAL: Program-aided Language Models. ICML 2023, 10764--10799. arXiv:2211.10435.\par\smallskip
Geng, Y., Li, H., Mu, H., Han, X., Baldwin, T., Abend, O., Hovy, E., \& Frermann, L. (2025). Control Illusion: The Failure of Instruction Hierarchies in Large Language Models. arXiv:2502.15851.\par\smallskip
Golikov, P., Opryshko, E., Pekhimenko, G., \& Jeffrey, M. C. (2026). Robust Reasoning Benchmark. arXiv:2604.08571.\par\smallskip
Hu, S., Luo, Y., Wang, H., Cheng, X., Liu, Z., \& Sun, M. (2023). Won't Get Fooled Again: Answering Questions with False Premises. ACL 2023, 5626--5643. arXiv:2307.02394.\par\smallskip
Huang, K., Guo, J., Li, Z., Ji, X., Ge, J., Li, W., Guo, Y., Cai, T., Yuan, H., Wang, R., Wu, Y., Yin, M., Tang, S., Huang, Y., Jin, C., Chen, X., Zhang, C., \& Wang, M. (2025). MATH-Perturb: Benchmarking LLMs' Math Reasoning Abilities against Hard Perturbations. ICML 2025. arXiv:2502.06453.\par\smallskip
Kadavath, S., Conerly, T., Askell, A., Henighan, T., Drain, D., Perez, E., Schiefer, N., Hatfield-Dodds, Z., et al. (2022). Language Models (Mostly) Know What They Know. arXiv:2207.05221.\par\smallskip
Levy, M., Jacoby, A., \& Goldberg, Y. (2024). Same Task, More Tokens: the Impact of Input Length on the Reasoning Performance of Large Language Models. ACL 2024, 15339--15353. arXiv:2402.14848.\par\smallskip
Li, Q., Cui, L., Zhao, X., Kong, L., \& Bi, W. (2024). GSM-Plus: A Comprehensive Benchmark for Evaluating the Robustness of LLMs as Mathematical Problem Solvers. ACL 2024, 2961--2984. arXiv:2402.19255.\par\smallskip
Liang, P., Bommasani, R., Lee, T., Tsipras, D., Soylu, D., Yasunaga, M., Zhang, Y., Narayanan, D., et al. (2022). Holistic Evaluation of Language Models. arXiv:2211.09110.\par\smallskip
Lin, S., Hilton, J., \& Evans, O. (2022). TruthfulQA: Measuring How Models Mimic Human Falsehoods. ACL 2022, 3214--3252. arXiv:2109.07958.\par\smallskip
Liu, N. F., Lin, K., Hewitt, J., Paranjape, A., Bevilacqua, M., Petroni, F., \& Liang, P. (2024). Lost in the Middle: How Language Models Use Long Contexts. TACL, 12, 157--173. arXiv:2307.03172.\par\smallskip
Lu, Y., Bartolo, M., Moore, A., Riedel, S., \& Stenetorp, P. (2022). Fantastically Ordered Prompts and Where to Find Them: Overcoming Few-Shot Prompt Order Sensitivity. ACL 2022, 8086--8098. arXiv:2104.08786.\par\smallskip
Luo, X., Jiang, X., \& Wu, J. (2026). MedDialBench: Benchmarking LLM Diagnostic Robustness under Parametric Adversarial Patient Behaviors. arXiv:2604.06846.\par\smallskip
Mirzadeh, I., Alizadeh, K., Shahrokhi, H., Tuzel, O., Bengio, S., \& Farajtabar, M. (2024). GSM-Symbolic: Understanding the Limitations of Mathematical Reasoning in Large Language Models. arXiv:2410.05229.\par\smallskip
Mizrahi, M., Kaplan, G., Malkin, D., Dror, R., Shahaf, D., \& Stanovsky, G. (2024). State of What Art? A Call for Multi-Prompt LLM Evaluation. TACL, 12, 933--949. arXiv:2401.00595.\par\smallskip
Nalbandyan, G., Shahbazyan, R., \& Bakhturina, E. (2025). SCORE: Systematic Consistency and Robustness Evaluation for Large Language Models. NAACL 2025 (Industry Track). arXiv:2503.00137.\par\smallskip
Rajpurkar, P., Jia, R., \& Liang, P. (2018). Know What You Don't Know: Unanswerable Questions for SQuAD. ACL 2018, 784--789. arXiv:1806.03822.\par\smallskip
Ribeiro, M. T., Singh, S., \& Guestrin, C. (2018). Semantically Equivalent Adversarial Rules for Debugging NLP Models. ACL 2018, 856--865.\par\smallskip
Ribeiro, M. T., Wu, T., Guestrin, C., \& Singh, S. (2020). Beyond Accuracy: Behavioral Testing of NLP Models with CheckList. ACL 2020, 4902--4912. arXiv:2005.04118.\par\smallskip
Sclar, M., Choi, Y., Tsvetkov, Y., \& Suhr, A. (2024). Quantifying Language Models' Sensitivity to Spurious Features in Prompt Design, or: How I learned to start worrying about prompt formatting. ICLR 2024. arXiv:2310.11324.\par\smallskip
Sharma, M., Tong, M., Korbak, T., Duvenaud, D., Askell, A., Bowman, S. R., Cheng, N., Durmus, E., et al. (2023). Towards Understanding Sycophancy in Language Models. arXiv:2310.13548.\par\smallskip
Shi, F., Chen, X., Misra, K., Scales, N., Dohan, D., Chi, E. H., Schärli, N., \& Zhou, D. (2023). Large Language Models Can Be Easily Distracted by Irrelevant Context. ICML 2023, 31210--31227. arXiv:2302.00093.\par\smallskip
Sun, Z., Dai, G., Tsang, I., \& Ye, H. (2025). Numerical Sensitivity and Robustness: Exploring the Flaws of Mathematical Reasoning in Large Language Models. arXiv:2511.08022.\par\smallskip
Wallace, E., Xiao, K., Leike, R., Weng, L., Heidecke, J., \& Beutel, A. (2024). The Instruction Hierarchy: Training LLMs to Prioritize Privileged Instructions. arXiv:2404.13208.\par\smallskip
Wang, J., Hu, X., Hou, W., Chen, H., Zheng, R., Wang, Y., Yang, L., Huang, H., Ye, W., Geng, X., Jiao, B., Zhang, Y., \& Xie, X. (2023). On the Robustness of ChatGPT: An Adversarial and Out-of-distribution Perspective. arXiv:2302.12095.\par\smallskip
Wang, Y., \& Zhao, Y. (2024). RUPBench: Benchmarking Reasoning Under Perturbations for Robustness Evaluation in Large Language Models. arXiv:2406.11020.\par\smallskip
Wu, Z., Qiu, L., Ross, A., Akyürek, E., Chen, B., Wang, B., Kim, N., Andreas, J., \& Kim, Y. (2024). Reasoning or Reciting? Exploring the Capabilities and Limitations of Language Models Through Counterfactual Tasks. NAACL 2024, 1819--1862. arXiv:2307.02477.\par\smallskip
Zhang, H., Da, J., Lee, D., Robinson, V., Wu, C., Song, W., Zhao, T., Raja, P., Zhuang, C., Slack, D., Lyu, Q., Hendryx, S., Kaplan, R., Lunati, M., \& Yue, S. (2024). A Careful Examination of Large Language Model Performance on Grade School Arithmetic. arXiv:2405.00332.\par\smallskip
Zhao, Z., Wallace, E., Feng, S., Klein, D., \& Singh, S. (2021). Calibrate Before Use: Improving Few-Shot Performance of Language Models. ICML 2021, 12697--12706. arXiv:2102.09690.\par\smallskip
Zhu, K., Wang, J., Zhou, J., Wang, Z., Chen, H., Wang, Y., Yang, L., Ye, W., Zhang, Y., Gong, N. Z., \& Xie, X. (2023). PromptRobust (PromptBench): Towards Evaluating the Robustness of Large Language Models on Adversarial Prompts. arXiv:2306.04528.\par\smallskip
\endgroup

\section*{Appendix A. Prompt templates}
\textbf{Answer format --- o4-mini, gemini-2.5-flash, gpt-4o-mini}

\begin{quote}\small\ttfamily
You are a math problem solver. Given a grade school math question, solve it step by step and provide the final numerical answer.\\
IMPORTANT: Your response must be ONLY a JSON object with this exact format:\\
\{"ground\_truth": \textless{}number\textgreater{}\}\\
Where \textless{}number\textgreater{} is the final numerical answer (integer or decimal, no units, no dollar signs, no commas).
\end{quote}

\textbf{Answer format --- phi4-mini}

\begin{quote}\small\ttfamily
You are a math problem solver. Solve the grade-school math problem step by step. Then give the final answer on its own line in exactly this format:\\
\#\#\#\# \textless{}number\textgreater{}\\
(a plain number only --- no units, \$, commas, or words).
\end{quote}

\textbf{Knowledge Boundary --- all models}

\begin{quote}\small\ttfamily
Solve the grade-school math problem step by step. If the question has a false premise, an impossible or fictional scenario, made-up entities, or is missing information needed to solve it, do NOT invent an answer, briefly explain why it cannot be answered. Otherwise end with the final answer on its own line as a plain number only (no units, \$, words, or commas):\\
\#\#\#\# \textless{}number\textgreater{}
\end{quote}

\end{document}